\pdfoutput=1
\documentclass[11pt]{article}
\usepackage{multirow}
\usepackage{acl}
\usepackage{authblk} % 使用authblk包
\usepackage{times}
\usepackage{latexsym}

\usepackage[T1]{fontenc}
\usepackage[utf8]{inputenc}

\usepackage{microtype}

\usepackage{amsmath}

\usepackage{inconsolata}
\usepackage{graphicx}
\usepackage{fancyhdr}
\fancypagestyle{plain}{
  \fancyhf{}
  \fancyfoot[C]{\thepage}
}

\title{MoE-CT: A Novel Approach For Large Language Models Training With Resistance To Catastrophic Forgetting}

\author[1]{Tianhao Li}
\author[2]{Shangjie Li}
\author[3]{Binbin Xie}
\author[2]{Deyi Xiong}
\author[1]{Baosong Yang\, \thanks{*  Corresponding author}\;   }

\affil[ ]{\textsuperscript{1}Alibaba Group \; \textsuperscript{2}Tianjin University \; \textsuperscript{3}Xiamen University}

\affil[ ]{
    \texttt{\{chongsheng.lth,\, yangbaosong.ybs\}}@alibaba-inc.com
}
\affil[ ]{\texttt{\{sj\_li, dyxiong\}}@tju.edu.cn,\; \texttt{xdblb}@stu.xmu.edu.cn}
\begin{document}
\maketitle
\thispagestyle{plain}  % Make sure the first page has page number

\begin{abstract}
The advent of large language models (LLMs) has predominantly catered to high-resource languages, leaving a disparity in performance for low-resource languages. Conventional Continual Training (CT) approaches to bridge this gap often undermine a model's original linguistic proficiency when expanding to multilingual contexts. Addressing this issue, we introduce a novel MoE-CT architecture, a paradigm that innovatively separates the base model's learning from the multilingual expansion process. Our design freezes the original LLM parameters, thus safeguarding its performance in high-resource languages, while an appended MoE module, trained on diverse language datasets, augments low-resource language proficiency. Our approach significantly outperforms conventional CT methods, as evidenced by our experiments, which show marked improvements in multilingual benchmarks without sacrificing the model's original language performance. Moreover, our MoE-CT framework demonstrates enhanced resistance to forgetting and superior transfer learning capabilities. By preserving the base model's integrity and focusing on strategic parameter expansion, our methodology advances multilingual language modeling and represents a significant step forward for low-resource language inclusion in LLMs, indicating a fruitful direction for future research in language technologies.

\end{abstract}

\section{Introduction}

Large language models (LLMs) \cite{ouyang2022training,DBLP:gpt3} have achieved remarkable progress in recent years, particularly in areas such as language generation \cite{Radford2019LanguageMA,NEURIPS2020_1457c0d6}, machine translation \cite{DBLP:journals/corr/VaswaniSPUJGKP17,wu2019google}, text summarization \cite{46111,rush-etal-2015-neural}, and language understanding \cite{Devlin2019BERTPO,peters-etal-2018-deep}. However, the majority of these models have focused on resource-rich languages such as English, leaving substantial potential for performance improvements in low-resource languages. To alleviate above mentioned issues, researchers have aimed to Parameter-Efficient Fine-Tuning (PEFT) methods. Specifically, continual training (CT) has been proposed, which have proven to be effective in enhancing the performance of low-resource languages by training on specific language data. Besides, other researchers have resorted to method of Low-Rank Adaptation (LoRA) \cite{hu2021lora}. By introducing a low-rank structure to reduce the model parameters for efficient updating. Thereby, it achieves a successful success between maintaining model performance and saving computation abd storage.
\begin{figure}
    \centering
    \includegraphics[width=1\linewidth]{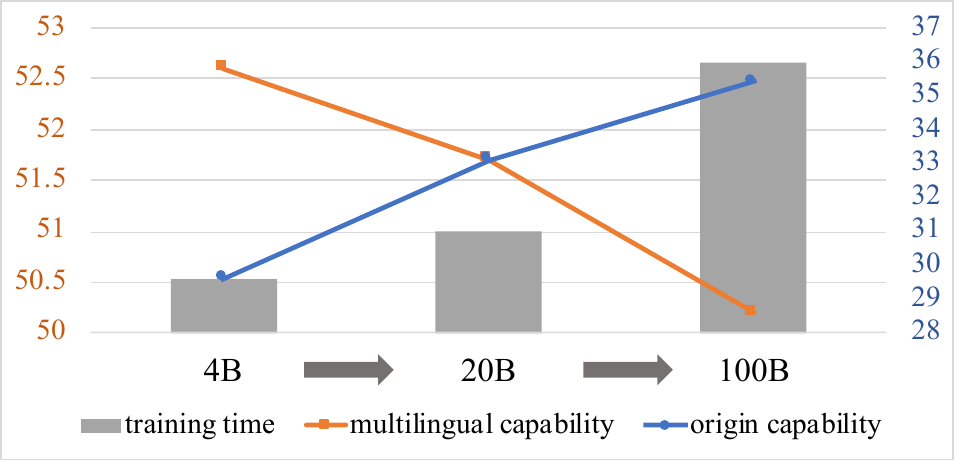}
    \caption{The abscissa represents the number of tokens from the original data incorporated during the Continual Training (CT) process. Although an increased volume of original data may decelerate the model's forgetting, it can significantly impede the enhancement of multilingual capabilities.}
    \label{fig2}
\end{figure}
% Despite these developments, studies on expanding the multilingual capabilities of large models are still limited. We conducted an in-depth analysis of existing capability extension technologies. Firstly, due to the difficulty in obtaining original training data, subsequent developers are virtually unable to utilize it in the customization process of models. As indicated in Table ~\ref{ratio}, the final performance of the models varies greatly when different proportions of mixed data are used during CT, it shows that the original data plays a key role in the knowledge stored later in the model; therefore, the absence of original data can lead to severe catastrophic forgetting. 
Despite these developments, research on how to extend the multilingual capabilities of large models are still limited. We conducted an in-depth analysis of existing capability extension technologies. 
\begin{enumerate}
    \item Firstly, \textbf{it is difficult to obtain original training data.} In conventional training methods, the original training data accounts for a significant proportion, so that the lack of data poses a significant challenge for the training in use. In addition, due to the different distributions between the training data of different stages, the absence of original training data will significantly exacerbate the catastrophic forgetting problems.
    % Furthermore, our experimental results, as shown in Table ~\ref{ratio}, indicate that in the traditional CT process, to prevent the collapse of the original language capabilities, the volume of original language data must be at least five times greater than that of the multilingual data being added. If the original data volume is substantially less than the multilingual data volume, then the model's original abilities will severely deteriorate. Thus, different proportions of mixed data used during CT yield vastly different outcomes, underscoring the critical role of original data in the knowledge preserved by the models in subsequent stages. In other words, the absence of original data can lead to serious catastrophic forgetting issues.

    \item Secondly, \textbf{adding large amounts of raw data will limit the improvement of multilingual capabilities.} As illustrated in Figure~\ref{fig2}, with the volume of original language data greatly alleviated the issue of catastrophic forgetting, significantly improving the generation capability in original data distribution. When the volume of original language data is five times greater than oth new one, the catastrophic forgetting issue is almost disappeared. However, it limits the model's final performance on multilingual tasks. Furthermore, a large amount of original data significantly increases the training costs of the model. 
    % Consequently, in the traditional CT process, the balance between the proportion of original data versus multilingual data, as well as the associated training costs, becomes a challenging issue to navigate for subsequent developers.
\end{enumerate}
In response to the aforementioned issues, we propose a method that utilizes a Mixture of Experts (MoE) \cite{DBLP:switch,DBLP:Gshard}approach. The Mixture of Experts model integrates multiple experts into the model architecture, where each expert is tasked with learning a specific task or feature subspace, thereby enhancing the model's learning capacity and generalization performance. Our specific approach involves extending additional expert networks on top of the pre-trained model to more efficiently learn new multilingual capabilities. To effectively prevent catastrophic forgetting, we froze most of the parameters of the original LLM. In addition, we employed a frozen shared feed-forward network (shared-ffn) to preserve the original knowledge, and implemented a gating mechanism to dynamically merge the original knowledge with the newly acquired knowledge from the expert networks. The crux of this method is that it enables us to effectively retain the capabilities of the base model, even with only a limited amount of original data. Our approach loosens the restrictions on the proportion of multilingual data, allowing it to play a more significant role during training. This not only raises the ceiling for the model's performance on multilingual tasks but also reduces training costs due to a substantial reduction in the overall volume of data.

In our proposed method, we selected Qwen as our foundational model and, based on this, expanded the MoE network with 2-8 experts per layer. We extracted Chinese and English data, as well as a substantial quantity of multilingual data, from Wikipedia and mC4 to serve as the continued training data. Our multilingual capability enhancements were validated using datasets from the xwinograd \cite{xwinograd}, XCOPA \cite{xcopa}, and pasw-X \cite{pawsx} tasks designed for multilingual understanding, while the MMLU \cite{mmlu}and C-Eval \cite{huang2023ceval}datasets were employed to verify the model's retention of original capabilities. Empirical evidence demonstrates that our MoE expansion strategy outperforms conventional continual training (CT) and LoRA in terms of multilingual improvement. Moreover, in tests of original capabilities, our approach exhibits a notable increase in resistance to forgetting when compared to conventional CT. Our scheme has achieved commendable results across various sizes of the Qwen model series.

To sum up, our contributions are as follows:
\begin{itemize}
    \item Our method introduces a novel MoE training paradigm for multilingual large model training, eliminating the need for extensive pre-training data while achieving compatibility between original and multilingual capabilities.
\end{itemize}

\begin{itemize}
    \item We employed additional expert networks to learn multilingual competencies, and by increasing the proportion of multilingual data, we have surpassed the upper limits of multilingual capabilities inherent in conventional CT methods.
\end{itemize}

\begin{itemize}
    \item Our approach leads the way in enhancing multilingual abilities and resistance to forgetting when compared to standard CT, demonstrating generalizability and transferability across models of varying scales.
\end{itemize}
% To sum up, our contributions are as follows:
% \begin{itemize}
%     \item Our approach, upon expanding large models with the target language capabilities, demonstrates superior extension performance compared to conventional continued training methods.
% \end{itemize}

% \begin{itemize}
%     \item Our training methodology has mitigated the issue of catastrophic forgetting to a certain extent during the continued training process of large models.
% \end{itemize}

% \begin{itemize}
%     \item Our Mixture of Experts (MoE) training approach, distinct from traditional continued training, does not require extensive pre-training data support and can achieve significant resistance to forgetting with only a minimal amount of pre-training data, thereby substantially reducing training costs.
% \end{itemize}

\begin{figure*}
\centering
\includegraphics[width=1\textwidth]{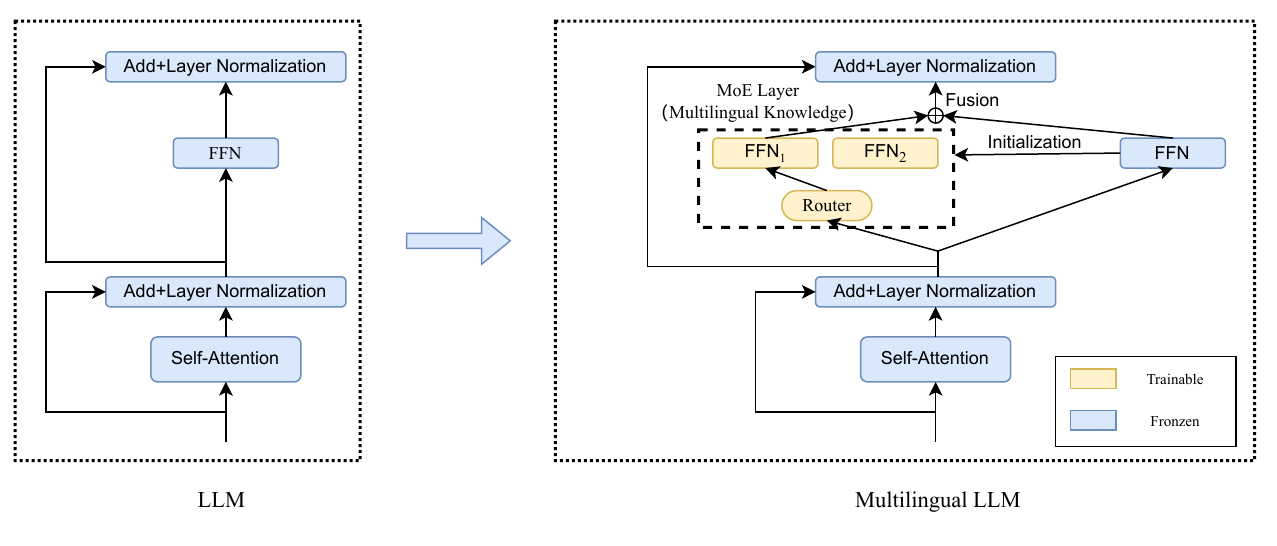}
\caption{The diagram on the right represents the training structure of MoE-CT, where the blue area indicates that parameters are frozen, and the yellow area indicates that parameters are trainable. The parameters for all experts and the shared feed-forward network (shared-ffn) are initialized from the feed-forward network (ffn) of the original model.}
\label{fig1}
\end{figure*}

\section{Related Work}

\textbf{Multilingual Pretraining and Fine-tuning in Language Models.}Developments in deep learning have significantly improved performance across a broad spectrum of natural language processing (NLP) tasks. Early efforts utilizing neural architectures such as recurrent neural networks (RNNs) \cite{DBLP:rnn}and long short-term memory (LSTM) \cite{sutskever2011generatinglstm} networks have laid the foundation for understanding and generating language representations \cite{DBLP:cho2014}. The advent of self-attention mechanisms and transformer architectures \cite{DBLP:journals/corr/VaswaniSPUJGKP17} has shifted the focus towards models that can be pretrained on large unlabeled corpora for better generalization across NLP tasks.Recent research has expanded the scope of these models to accommodate multiple languages, yielding multilingual models such as mBERT \cite{mbert} and XLM \cite{DBLP:xlm}, which are capable of learning cross-lingual representations. By pretraining on text from various languages, these models leverage shared linguistic properties and can be fine-tuned for downstream tasks in different languages.

\textbf{Sparsely Gated Networks} Efficiency in neural network scaling has gained paramount importance as the size of state-of-the-art language models has ballooned, resulting in steep computational costs. Sparsely gated networks have emerged as a promising direction for realizing model scaling without a linear increase in resource demands. At the core of this approach lies the concept of conditional computation, where different parts of the network are activated based on the input. The pioneering work of \cite{DBLP:journals/corr/ShazeerMMDLHD17} demonstrated the potential of sparsely gated mixture-of-experts (MoE) layers in language models, where only a subset of experts is chosen for each input, drastically reducing computation during inference. This is achieved through gating mechanisms that learn to distribute the computation across a diverse set of experts, each specializing in different aspects of the data. Concurrently, a growing body of work is focusing on extending the Mixture of Experts (MoE) architecture with sparse activations \cite{DBLP:journals/corr/Hestness,DBLP:journals/corr/Shazeer2018,kudugunta2021beyond}.Switch Transformer \cite{DBLP:switch} and GLaM \cite{DBLP:journals/corr/abs-2112-06905}showcased a model with orders of magnitude more parameters than traditional models, but with fewer activated parameters per example, leading to improved performance and training efficiency. Similarly, the GShard framework \cite{DBLP:Gshard} demonstrated the scalability of MoE for large-scale multilingual machine translation tasks, reinforcing the viability of sparse gating for extensive language coverage.

% \textbf{Continual Learning for Large Language models} Continual learning, also referred to as lifelong learning, addresses the challenge of creating large language models that can learn from a continuous stream of information without forgetting previously acquired knowledge. This is particularly relevant for language models which must adapt to evolving data distributions and new tasks over time without losing proficiency in earlier domains 

\textbf{Continual Learning for Large Language models} Advancements in Natural Language Processing (NLP) have led to the exploration of various strategies to support continual learning capabilities. These strategies encompass: i) techniques that leverage the concept of replay to retain knowledge  \cite{robins,DBLP:journals/corr/RebuffiKL16,DBLP:journals/corr/ShinLKK17,DBLP:journals/corr/Lopez-PazR17,DBLP:journals/corr/chaudhry2018}; ii) strategies grounded in regularization to prevent overwriting of existing information \cite{DBLP:journals/corr/KirkpatrickPRVD16,DBLP:journals/corr/LiHoiem}; and iii) designs that restructure the neural architecture itself to accommodate new learning\cite{DBLP:journals/corr/RusuRDSKKPH16,DBLP:journals/corr/yoon,DBLP:journals/corr/mallya,DBLP:journals/corr/yemingwen} . The domain of NLP has witnessed a burgeoning interest in these continual learning paradigms, as evidenced by a series of recent contributions\cite{wang-etal-2019-sentence,biesialska-etal-2020-continual,sun2019lamol,huang-etal-2021-continual,Hussain2021TowardsAR,DBLP:journals/corr/arhens2021,Jin2021LifelongLO,Lin2022OnCM}. These include embedding alignment with episodic memory \cite{wang2019}; improvements to memory-based parameter adaptation via sparse experience replay \cite{autume2019}; approaches to lifelong language modeling \cite{Sun2019LAMOLLM}; and the adoption of meta-learning frameworks that integrate sparse replay \cite{Holla2020MetaLearningWS}. \cite{lifelong} marks the first application of the Mixture of Experts (MoE) structure to expand NLP tasks, mitigating the issue of catastrophic forgetting encountered in cross-task learning. Traditional approaches have predominantly focused on models of small to medium scale, where recent research has made strides in mitigating catastrophic forgetting to a certain degree. However, these strategies are often not applicable to large models with substantial parameter counts. In the continued training phase, such large models typically confront a more pronounced issue of catastrophic forgetting. Fine-tuning these models to adapt to specific tasks can lead to rapid loss of previously acquired general knowledge, even with minor parameter updates.In light of this, our work is dedicated to exploring how to balance the trade-off between stability (retaining old knowledge) and plasticity (acquiring new knowledge). By investigating and developing strategies that enable large models to maintain previously learned information while also learning new tasks effectively, we aim to contribute to the field's understanding of how to construct artificial intelligence systems capable of continual learning without sacrificing their pre-trained knowledge.
% A critical concern within the domain of lifelong learning is addressing catastrophic forgetting. Notwithstanding, existing research often concentrates on sequential task learning, neglecting the real-world applicability of pre-trained language models that can swiftly adapt to new tasks with minimal data. Such task-level lifelong learning does not cater to the typical requirements of operational NLP applications, where model updates occur incrementally. By contrast, our work directs its focus toward the pretraining of language models under a continuous influx of evolving data distributions—data-level lifelong pretraining—which more accurately mirrors the practical needs for ongoing deployment and refinement of NLP systems.

\begin{table*}
    \centering
    \begin{tabular}{c|cccccccccccc}
    \hline
         & en & zh & ar & id & vi & de & fr & ja & th & pt & es & it \\ \hline
      Tokens   & 2B & 2B & 2B & 2B & 2B & 2B & 2B & 2B & 2B & 2B & 2B & 2B \\ \hline
    \end{tabular}
    \caption{The composition of the continual training dataset, there are 24B tokens in total, and each language is allocated an equal proportion of the data.}
    \label{tokens}
\end{table*}

\begin{table*}
    \centering
    \begin{tabular}{ccccccc}
    \hline
        \textit{Type} & \textit{Expert} & \textit{Params} & \textit{Act-params} & \textit{Layers} & \textit{Hidden size} & \textit{Heads} \\ \hline
        Dense & - & 1.8B & 1.8B & 24 & 2048 & 16\\
        MoE & 2 & 3.4B & 2.6B & 24 & 2048 &  16 \\ \hline
        Dense & - & 7B & 7B & 32 & 4096 & 32  \\
        MoE & $2\sim8$ & $16B\sim40B$ & 11B & 32 & 4096 & 32  \\
        \hline
    \end{tabular}
    \caption{The size and  architecture of our MoE model and dense model}
    \label{params}
\end{table*}

\section{Method}
% In response to the substantial integration of multilingual knowledge, there is a necessity to increase the parameter count of the model to provide sufficient support. Concurrently, to preserve the inherent abilities of the model, we strive to minimize alterations to the original parameter distribution. In light of these considerations, we have devised an approach for expanding large language models through a Mixture of Experts (MoE) methodology, with parameter modifications predominantly concentrated in the experts and routing mechanism. To elaborate: 1) We have incorporated additional expert networks at each layer of the model, which are specifically tailored to learn and amplify the model's extended multilingual capabilities. 2) With the intent to retain the model's original functionality, we have kept all parameters static, barring those associated with the experts and embedding layers. By employing this strategy and through the training of the routing mechanism, we guide the model towards an optimal equilibrium between its augmented faculties and original competencies.
As discussed above, it is evident that the multilingual capabilities of large-scale models remain constrained.Our goal is to maintain a robust performance in the generation of high-resource languages, effectively mitigating the risk of catastrophic forgetting, while simultaneously enhancing the multilingual capabilities of the model.

To achieve this, we introduce a novel architectural paradigm named MoE-CT, which substantially integrates diverse knowledge domains into our model, by employing an \textsc{ MoE} (Mixture of Experts) approach tailored specifically for the multilingual context. 
\subsection{Mixture of Experts}
In the described model, we have a collection of N feed-forward neural networks, which are all structurally equivalent and operate independently, denoted as a set of experts ${E_i}{i=1}^N$. These experts are integrated with a dedicated gating mechanism, denoted as $G(\cdot)$, which functions as a decision-maker. This gating mechanism is responsible for determining the contribution of each expert's output in the final response of the system. Specifically, if we consider $h$ to be the resultant vector from an attention mechanism in any block of the model, then the final output $y$ from the mixture of experts (MoE) layer is derived by taking a weighted sum of each expert's output. Mathematically, this can be described by the equation:
$$
y = \sum{i=1}^N G(h)_i \cdot E_i(h)
$$
In this expression, $E_i(h)$ represents the output of the $i$-th expert network, while $G(h)_i$ signifies the weight allocated to that output by the gating function. The gating function itself is defined using a softmax operation applied to the dot product of the input $h$ and a set of learned parameters encapsulated in the matrix $W_g$, which is expressed as:
$$
G(\cdot) = \text{Softmax}(h \cdot W_g)
$$
In essence, the matrix $W_g$ holds the parameters that the gating function adapts during training to optimize the routing of information through the various experts in the MoE architecture.
\subsection{Routing mechanism design}
Our goal is to store multilingual knowledge in the MoE layer and store English/Chinese knowledge in the old FFN layer of pretrained LLM. We hope that the sparse extended multilingual LLM can combine the outputs of the two layers to achieve the best world of multilingual and English/Chinese abilities. Therefore, we have added a fusion module with different structures to explore the best combination of MoE and FFN layers. The output of the fusion module can be described as follows:
\begin{equation}
    \text{Fusion}(x, y) = w \cdot x + (1 - w) \cdot y
\end{equation}
In the equation, $w$ represents the weight of the MoE layer, which, when combined with the weight of the shared-ffn, sums to 1.

\begin{equation}
    \text{output}=Fusion\{\sum_{i=1}^{K} w_i \text{FFN}_i(h_{in}), \text{FFN}(h_{in})\}
\end{equation}
where $w_i$ is the weight of $i$-th expert determined by router, and $K$ is the Top-$K$ experts selected by the router.

To prevent the problem of \textit{catastrophic forgetting} during multilingual training, most of the parameters in our sparse-extended multilingual LLM will be fronzen, and only the MoE layer, embedding layer and fusion module are trainable, as shown in Figure~\ref{fig1}, the blue modules are frozen, and the yellow modules are trainable.

\subsection{An innovative model training methodology}
% We leverage Qwen\cite{bai2023qwen} , a large language model pre-trained using a large amount of Chinese and English data
% as our base model, exhibits industry-leading performance in both Chinese and English. Due to limitations in training resources, we employ the 1b8 and 7b versions of Qwen's models as our foundational architectures.
% sparse structures by adding MoE layer \cite{DBLP:journals/corr/ShazeerMMDLHD17,DBLP:switch}parallel to the Feed Forward Layer. The MoE layer includes multiple FFN layers, but only a small portion will be activated, and the activation strategy is determined by the router module. We use the parameters of FFN layer from pretrained LLM to initialize multiple FFN layers in the MoE layer, which can significantly accelerate the training process and provide valuable knowledge transfer from English/Chinese to other languages. 
As shown in Figure~\ref{fig1}, our method extends large language models from dense structures to sparse structures by adding MoE layer parallel to the Feed Forward Layer. The MoE layer includes multiple FFN layers, but only a small portion will be activated, and the activation strategy is determined by the router module. We use the parameters of FFN layer from pretrained LLM to initialize multiple FFN layers in the MoE layer, which can significantly accelerate the training process and provide valuable knowledge transfer from old one to new one.. 

Throughout the continuation training process, we exclusively train the expert networks and the embedding layer within the expanded MoE model architecture, while all other structural parameters and shared-FFN parameters are kept fixed. Through experimental validation, we have ascertained that such a training strategy optimally retains the model's original capabilities while effectively expanding its multilingual capacity.
\section{Experiment}


% \usepackage{multirow}

\begin{table*}[]
\centering
\resizebox{1.8\columnwidth}{!}{
\begin{tabular}{llrrrl}
\hline
Task category              & Task   & Test      & Lang. & Metric & Prompt                                                  \\ \hline
NLU       & XNLI   & 5,010     & 4     & Acc.   & [Premise], right? \{Yes/Also/No\}, [Hypothesis]           \\
                           & XCOPA  & 500       & 7     & Acc.   & [Prefix] \{because/therefore\} \{choice1/choice2\} [Suffix] \\
                           & PAWS-X & 2,000     & 7     & Acc.   & [Sentence1], right? \{Yes/No\}, [Sentence2]               \\ \hline
Knowledge & C-Eval & 1590         & 1    & Acc.   & [Question]\{Choices\}[Answer]                             \\
                           & MMLU   & 14213         & 1    & Acc.   & [Question]\{Choices\}[Answer]                             \\ \hline
MT                         & WMT/IWSLT  & 991-3,002 & 9     & BLEU   & [INPUT] Translate this sentence from [SRC] to [TGT].    \\ \hline
\end{tabular}}

\caption{Multilingual benchmark}
\label{table_bench}
\end{table*}

% \begin{table*}[]
% \centering
% \resizebox{1.8\columnwidth}{!}{
% \begin{tabular}{c|ccc|ccc}
% \hline
% {Model}      & \multicolumn{3}{c|}{Tokens}                                             & \multicolumn{1}{c|}{Chinese Ability} & \multicolumn{1}{c|}{English Ability} & Multilingual Ability \\ \cline{2-7} 
%  & EN & ZH & Multilingual & C-eval & MMLU & 3on1 \\ \hline
% Qwen-1b8 & - & - & - & 48.5 & 37 & 48.6\\
% Qwen-1b8-normal-CT & 50B & 50B & 20B & 39.5 & 31.3 & 50.2\\
% Qwen-1b8-uniform-CT & 2B & 2B & 20B & 29.7 & 29.5 & 52.6\\
% Qwen-1b8-uniform-CT-MoE & 2B & 2B & 20B & \textbf{41.5} & \textbf{33.4} & \textbf{53.3}\\
% Qwen-7b & - & - & - & 57.4 & 42.7 & 54.4\\
% Qwen-7b-normal-CT & 50B & 50B & 20B & 53.3 & 41.1 & 57.6\\
% Qwen-7b-uniform-CT & 2B & 2B & 20B & 50.1 & 39.2 & 58.9\\
% Qwen-7b-uniform-CT-MoE & 2B & 2B & 20B & \textbf{55.5} & \textbf{42.3} & \textbf{58.9}\\ \hline
% \end{tabular}}
% \caption{The impact of data matching}
% \label{ratio}
% \end{table*}

\begin{table*}[]
\centering
\resizebox{2.1\columnwidth}{!}{
\begin{tabular}{c|ccccccc|cc}
\hline
{Model}      & \multicolumn{7}{c|}{Multilingual Ability}                                             & \multicolumn{1}{c|}{Chinese Ability} & English Ability \\ \cline{2-10} 
                            & XCOPA & PAWS-X & XNLI & MT (en2xx) & MT (xx2en) & MT (en2zh) & MT (zh2en) & C-eval          & MMLU            \\ \hline
Qwen-1b8                    & 54.7  & 53.0     & 38   & 15.1          & 21.0            & 35.6          & 22.1          & 48.5                                 & 37.0              \\
Qwen-1b8-CT                 & 62.3  & 51.0     & 44.7 & 19.1          & 20.9          & 27.2          & 15.2          & 29.7                                 & 29.5            \\
Qwen-1b8-MoE-CT (Experts=1) & 62.4  & 51.7   & 44.2 & 19.0            & 21.1          & 29.7          & 17.8          & 40.5                                 & 32.2            \\
Qwen-1b8-MoE-CT (Experts=2) & 62.5  & 52.9   & 44.4 & 19.5          & 21.9          & 30.8          & 18.5          & 41.5                                 & 33.4           \\ \hline
\end{tabular}}
\caption{Main results on Qwen-1b8.When compared to Qwen-CT, the Qwen-MoE-CT exhibits a more robust resistance to forgetting and also demonstrates a greater enhancement in multilingual capabilities.}
\label{table1}
\end{table*}

% \usepackage{multirow}
\begin{table*}[]
\centering
\resizebox{2.1\columnwidth}{!}{
\begin{tabular}{c|ccccccc|cc}
\hline
Model     & \multicolumn{7}{c|}{Multilingual Ability}                                             & \multicolumn{1}{c|}{Chinese Ability} & English Ability \\ \cline{2-10} 
                           & XCOPA & PAWS-X & XNLI & MT (en2xx) & MT (xx2en) & MT (en2zh) & MT (zh2en) & C-eval                               & MMLU            \\ \hline
Qwen-7b                    & 61.6  & 60.1   & 41.6 & 24.0          & 30.2          & 40.1          & 29.0            & 57.4                                 & 42.7            \\
Qwen-7b-CT                 & 69.8  & 60.1   & 46.8 & 27.5          & 30.3          & 37.7          & 26.8          & 50.1                                 & 39.2            \\
Qwen-7b-LoRA-CT               & 62.3  & 59.6   & 44.0   & 24.4          & 29.1          & 40.0            & 27.8          & 55.4                                 & 43.3            \\
Qwen-7b-MoE-CT (Experts=2) & 69.5  & 60.6   & 46.0   & 27.8          & 30.3          & 39.3          & 27.5          & 53.6                                 & 41.7            \\
Qwen-7b-MoE-CT (Experts=4) & 69.3  & 60.5   & 46.0   & 28.3          & 30.5          & 39.8          & 27.7          & 55.9                                 & 42.1            \\
Qwen-7b-MoE-CT (Experts=8) & 69.6  &   60.6     &  46.5    &        28.4       &    30.7           & 39.8          & 28.1          & 55.5                                 &         42.3        \\ \hline
\end{tabular}}
\caption{Main results on Qwen-7b.Our MoE architecture is equally applicable to various model sizes, and we have achieved substantial resistance to forgetting on Qwen-7b, alongside a comprehensive enhancement of multilingual capabilities that surpasses those of conventional CT and LoRA-CT.}
\label{table2}
\end{table*}

\subsection{Training Datasets}
The composition of our data and the proportion of each language variety are reflected in Table ~\ref{tokens}. Our continued training data set comprises a total of 24 billion tokens, of which 20 billion are multilingual data encompassing ten languages (Arabic, Indonesian, Vietnamese, German, French, Japanese, Thai, Portuguese, Spanish, and Italian), with each language accounting for 2 billion tokens. Additionally, there are 2 billion tokens of Chinese data and 2 billion tokens of English data. All of the data were extracted from mC4 \cite{mC4} and Wikipedia.

In addition, we prepared an extra 50 billion tokens for Chinese and 50 billion tokens for English to ensure that the proficiency in these languages does not diminish during the standard continual training (CT) process. Empirical evidence suggests that for the expanded multilingual dataset of 20 billion tokens, we need to incorporate at least five times more Chinese and English data to maintain their resistance to forgetting. However, within our Mixture of Experts (MoE) expanded architecture, we found that only 4 billion tokens of Chinese and English data are required to achieve resistance to forgetting in these languages, which substantially reduces the cost of training data.

\subsection{Architecture Setting}
We leverage Qwen \cite{bai2023qwen} , a large language model pre-trained using a large amount of Chinese and English data as our base model, exhibits industry-leading performance in both Chinese and English. Due to limitations in training resources, we employ the 1b8 and 7b versions of Qwen's models as our foundational architectures.

Table ~\ref{params} shows the parameter structure of Qwen as a dense model, as well as the parameter structure after extending Qwen based on MoE. "Expert" represents the number of experts utilized in the MoE architecture. "Params" refers to the total number of parameters in the model. "Act-params" denotes the quantity of parameters activated during model inference. "Layers" indicates the number of layers in the model. "Hidden size" is the hidden dimension of the feed-forward layers, and "Heads" is the number of self-attention heads.

\subsection{Multilingual Benchmark}
We used six test datasets to evaluate the multilingual and Chinese/English capabilities of our models, and we divided them into three task categories: \textbf{NLU}, \textbf{Knowledge task} and \textbf{Machine Translation}. Please refer to Table~\ref{table_bench} for detailed information.

\paragraph{NLU} For Natural Language Understanding (NLU) task category, we choose three multilingual test task: XNLI, XCOPA and PAWS-X.
\paragraph{Knowledge} The Knowledge task category includes two test dataset, C-Eval and MMLU, which are used to evaluate the Chinese and English capabilities of our models respectively.
\paragraph{MT} For Machine Translation tasks, we selected 9 language test datasets from WMT and IWSLT to evaluate the translation ability from these languages to English and English to these languages.
% \begin{table*}[]
% \centering
% \resizebox{1.8\columnwidth}{!}{
% \begin{tabular}{c|ccc|ccc}
% \hline
% {Model}      & \multicolumn{3}{c|}{Continue training Tokens}                                             & \multicolumn{1}{c|}{Chinese Ability} & \multicolumn{1}{c|}{English Ability} & Multilingual Ability \\ \cline{2-7} 
%  & EN & ZH & Multilingual & C-eval & MMLU & Avg \\ \hline
% Qwen-1b8 & - & - & - & 48.5 & 37 & 48.6\\
% Qwen-1b8-normal-CT & 50B & 50B & 20B & 39.5 & 31.3 & 50.2\\
% Qwen-1b8-uniform-CT & 2B & 2B & 20B & 29.7 & 29.5 & 52.6\\
% Qwen-1b8-uniform-CT-MoE & 2B & 2B & 20B & \textbf{41.5} & \textbf{33.4} & \textbf{53.3}\\
% Qwen-7b & - & - & - & 57.4 & 42.7 & 54.4\\
% Qwen-7b-normal-CT & 50B & 50B & 20B & 53.3 & 41.1 & 57.6\\
% Qwen-7b-uniform-CT & 2B & 2B & 20B & 50.1 & 39.2 & 58.9\\
% Qwen-7b-uniform-CT-MoE & 2B & 2B & 20B & \textbf{55.5} & \textbf{42.3} & \textbf{58.9}\\ \hline
% \end{tabular}}
% \caption{The data ratios required for conventional CT and those necessitated by the MoE expansion indicate that the MoE architecture can achieve better resistance to forgetting in Chinese and English, as well as enhanced multilingual capabilities, with only a minimal amount of Chinese and English data.}
% \label{ratio}
% \end{table*}

\subsection{Main Results}
The main results are shown in Table~\ref{table1} and Table~\ref{table2}, we conducted experiments on Qwen-1b8 and Qwen-7b models to verify the general effectiveness of our method. As shown in Table~\ref{table1}, Qwen-1b8-CT is a multilingual continue-training version of Qwen-1b8, which has a significant improvements of multilingual abilities over Qwen-1b8 model, improves the accuracy of XCOPA from 54.7\% to 62.3\%, XNLI from 38.0\% to 44.7\%. For machine translation task, multilingual continue-training brings in an average of 4.0 BLEU on En-XX directions, but no improvement is found on XX-EN directions. 

Although continue-training strategy can improve multilingual ability of LLM, it always causes problem of \textit{catastrophic forgetting}, which damages the original abilities of the model. We can find that Qwen-1b8-CT model has a significant drop on English-to-Chinese and Chinese-to-English translation performance, from 35.6 to 27.2 and 22.1 to 15.2, respectively. For Chinese and English ability evaluation, Qwen-1b8-CT achieves a performance decline from 48.5 to 29.7 and 37.0 to 29.5 on C-Eval and MMLU test set, respectively. What's more, simply continue train the LLM on multilingual corpus may not improve multilingual abilities, as we find a performance decline on PAWS-X test dataset (53.0 vs 51.0). \textit{Catastrophic forgetting} problem can also be found on Qwen-7b model in Table~\ref{table2}. We have also experimented with the use of LoRA-CT, setting the LoRA dimension to 8, as shown in Table ~\ref{table2}, although LoRA-CT can effectively alleviate the issue of catastrophic forgetting, its limited number of changed parameters results in a significant performance gap in extended multilingual capabilities when compared to conventional CT.

By using the sparse MoE architecture for continual training of LLM, the best results can be achieved in both multilingual and Chinese and English ability. In Table~\ref{table2}, we can find that Qwen-7b-MoE-CT performs equivalently to Qwen-7b-CT in multilingual tasks, and significantly prevents the performance degradation of Chinese and English capabilities, where performance on C-eval only decreases from 57.4 to 55.5 and MMLU decreases from 42.7 to 42.3. Our method also alleviates the decline in the performance of English-to-Chinese and Chinese-to-English translation, and has a significant improvement in the translation ability on English-XX directions from 24.0 to 28.4. The details of translation results on Qwen-1b8 and Qwen-7b can be found in Table~\ref{table3} and Table~\ref{table4}.
\begin{table*}[]
\centering
\resizebox{1.8\columnwidth}{!}{
\begin{tabular}{c|ccc|ccc}
\hline
{Model}      & \multicolumn{3}{c|}{Continue training Tokens}                                             & \multicolumn{1}{c|}{Chinese Ability} & \multicolumn{1}{c|}{English Ability} & Multilingual Ability \\ \cline{2-7} 
 & EN & ZH & Multilingual & C-eval & MMLU & Avg \\ \hline
Qwen-1b8 & - & - & - & 48.5 & 37 & 48.6\\
Qwen-1b8-CT & 50B & 50B & 20B & 39.5 & 31.3 & 50.2\\
Qwen-1b8-CT & 2B & 2B & 20B & 29.7 & 29.5 & 52.6\\
Qwen-1b8-MoE-CT & 2B & 2B & 20B & \textbf{41.5} & \textbf{33.4} & \textbf{53.3}\\
Qwen-7b & - & - & - & 57.4 & 42.7 & 54.4\\
Qwen-7b-CT & 50B & 50B & 20B & 53.3 & 41.1 & 57.6\\
Qwen-7b-CT & 2B & 2B & 20B & 50.1 & 39.2 & 58.9\\
Qwen-7b-MoE-CT & 2B & 2B & 20B & \textbf{55.5} & \textbf{42.3} & \textbf{58.9}\\ \hline
\end{tabular}}
\caption{The data ratios required for conventional CT and those necessitated by the MoE expansion indicate that the MoE architecture can achieve better resistance to forgetting in Chinese and English, as well as enhanced multilingual capabilities, with only a minimal amount of Chinese and English data.}
\label{ratio}
\end{table*}
In our experiments, we have identified a significant data ratio problem when employing conventional continual training (CT) methods to combat catastrophic forgetting. Specifically, to preserve the proficiency in Chinese and English, the volume of data for these languages must be at least five times greater than that of the multilingual dataset. However, as indicated in Table ~\ref{ratio}, such an excessive reliance on Chinese and English data significantly hampers the enhancement of multilingual capabilities. Moreover, it substantially increases the training cost, posing a major obstacle to the expansion of large language models.

In contrast to the aforesaid conventional CT, our proposed MoE expansion technique significantly reduces the dependence on Chinese and English data, with these languages' data requiring only a one-fifth proportion of the multilingual dataset. As demonstrated in Table ~\ref{ratio}, due to the reduced incorporation of Chinese and English data in our MoE framework, the model can more effectively assimilate multilingual knowledge. Consequently, the amplification of multilingual abilities is more pronounced compared to the conventional CT methods. This approach highlights the efficiency of our MoE expansion method in achieving a better and more efficient balance between preserving original language capabilities and enhancing expanded multilingual proficiencies.
\subsection{Ablation Study}
In this study, we conducted ablation experiments on the freezing strategy, routing mechanism, and the number of experts used in the MoE expanded architecture to verify their effects on multilingual improvement and resistance to forgetting of Chinese-English bilingual capabilities.

\begin{table}
    \centering
    \begin{tabular}{ccc}
    \hline
        Trainable parameters & original & expanded\\
        \small{Qwen-1b8} & 42.7 & 48.6 \\
        \small{all} & 30.8 & 52.8 \\
        \small{attention} & 31.3 & 51.5 \\
        \small{embedding} & 35.3 & 52.4 \\
        \small{experts} & 35.9 &  52.7\\
        \small{embedding\&experts} & 37.5 & 53.3 \\ \hline
    \end{tabular}
    \caption{The Impact of Training Strategies on Original and Expanded Capabilities in the MoE architecture, all means that all parameters participate in the training}
    \label{freeze}
\end{table}
\paragraph{Training strategy.}
% In our experiments on the Qwen-1.8B scale model, we attempted to freeze different parts of the MoE model to determine which components could retain the model's original capabilities. Specifically, we froze the embedding layer, the attention layer, and the shared feed-forward layer. As indicated in Table~\ref{freeze}, freezing the attention layer did not effectively mitigate the issue of forgetting. In contrast, freezing either the embedding layer or the feed-forward layer enhanced the model's resistance to forgetting. When both layers were frozen concurrently, the model's resistance to forgetting was further improved. Consequently, the final model structure incorporated a freezing of both the embedding layer and the shared ffn.
In our experiments on the Qwen-1.8B scale model, we attempted to freeze different parts of the MoE model to determine which components could retain the model’s original capabilities. As indicated in Table~\ref{freeze}, continual training all parameters or attention layers will result in a significant decrease in the model's original ability, while training expert layers and embedding layers can balance the original ability and expanded ability. Therefore, in the final model, we chose to train the parameters of the experts and the embedding layer. 
\paragraph{Routing mechanism.}
In our experiments, we attempted various combination methods for the outputs of the shared feed-forward network  and the MoE layer. Initially, we utilized a weighted sum approach for integrating the two, assigning MoE output weights from 0.1 to 0.9. Our experiments revealed that a lower MoE weight corresponded to a more pronounced resistance to catastrophic forgetting, yet resulted in a smaller improvement in multilingual capabilities. Conversely, a higher MoE weight weakened the resistance to catastrophic forgetting while yielding a greater enhancement in multilingual capabilities. As shown in table ~\ref{tab:table10}, when the weights for both the MoE and the shared-ffn are set to 0.5, an optimal balance is achieved between the enhancement of multilingual capabilities and the resistance to forgetting.

% The simple weighted sum approach proved inadequate for balancing between the two aspects. Consequently, we implemented a gating mechanism to dynamically balance the outputs of both components. As shown in table ~\ref{tab:table10}, the empirical results demonstrate that the gating mechanism successfully achieves an optimal balance between resistance to catastrophic forgetting and multilingual capabilities.
% \usepackage{multirow}
% \begin{table}
%     \centering
%     \begin{tabular}{cccc}
%     \hline
%          ~ & MoE weights & original & expanded\\ \hline
%          \multirow{2}{*}{Weighted} & 0.1 & 37.8 & 52.1 \\
%          ~ & 0.9 & 34.4 & 53.1 \\ \hline
%        Gating & Dynamic & 37.5 & 53.3 \\ 
%        \hline
%     \end{tabular}
%     \caption{The impact of different combinations of MoE layers and shared FFN on original and expanded capabilities in the Qwen-1B8 model, where MoE weight represents the proportion of the MoE layer's output in the total output.}
%     \label{tab:table10}
% \end{table}

\paragraph{The number of experts.}
 As shown in Table~\ref{table1} and Table~\ref{table2},  two experts already exhibits considerable resistance to forgetting of Chinese-English bilingual capabilities. To verify whether a greater number of experts would yield further enhancements in multilingual capabilities, we expanded the model to include 4 and 8 experts at the Qwen-7B scale. However, according to the experimental results, increasing the number of experts does not significantly improve multilingual abilities. In light of this, we hypothesize that the quantity of continued training data may be insufficient, preventing the multiple experts from converging to their optimal performance. Consequently, we plan to utilize a larger corpus of continued training data, provided sufficient training resources, to support an increased number of experts in future experiments.

\begin{table}
    \centering
    \begin{tabular}{cccc}
    \hline
         ~ & MoE weights & original & expanded\\ \hline
         \multirow{3}{*}{Weighted} & 0.1 & 37.8 & 52.1 \\
         ~ & 0.9 & 34.4 & 53.3 \\ 
       ~ & 0.5 & 37.5 & 53.3 \\ 
       \hline
    \end{tabular}
    \caption{The impact of different combinations of MoE layers and shared FFN on original and expanded capabilities in the Qwen-1B8 model, where MoE weight represents the proportion of the MoE layer's output in the total output.}
    \label{tab:table10}
\end{table}

\section{Conclusion}
In conclusion, our research presents a significant advancement in the field of multilingual language modeling, particularly addressing the challenges posed by catastrophic forgetting in large language models (LLMs). Through the introduction of the MoE-CT structure, we have demonstrated a novel approach that not only enhances the extension of LLMs to low-resource languages but also preserves the original linguistic competencies in high-resource languages. Our experiments on models such as Qwen-1b8 and Qwen-7b have validated the effectiveness of MoE-CT, marking clear improvements in multilingual benchmarks while maintaining or even improving performance in the original languages.

The MoE-CT structure showcases a delicate balance between stability and plasticity, ensuring that the base model's parameters remain undisturbed and that the newly introduced MoE layers absorb the additional linguistic knowledge. This balance is crucial for achieving a harmonious integration of multilingual capabilities without the detriment of pre-existing language proficiencies. Our findings indicate that MoE-CT can achieve substantial resistance to forgetting with a minimal amount of pre-training data, which is a considerable stride towards reducing training costs and resources.

The implications of our work are manifold. Primarily, it facilitates the creation of more inclusive language technologies that do not favor solely high-resource languages. Furthermore, it paves the way for future research into continual learning for LLMs, emphasizing the importance of models that can continually evolve and adapt to new language without losing previously established knowledge.

\section*{Limitations}
In summary, we have proposed the MoE-CT architecture to address the issue of catastrophic forgetting encountered by LLMs during the expansion of multilingual capabilities. Due to the limitations of computational resources, we have not attempted to extend the MoE-CT architecture to other open-source models beyond Qwen, which may not fully demonstrate the catastrophic forgetting challenges faced by all LLMs. Therefore, our future work will explore whether this structure can be adapted to a wider range of open-source models.

\section*{Ethics Statement}
Our work on the MoE-CT for LLMs considers several ethical concerns. Primarily, we aim to address linguistic biases by enhancing LLMs performance in low-resource languages, promoting inclusivity and cultural diversity. We recognize the risk of potential biases in model training and commit to their mitigation.
We also acknowledge the responsibility to prevent the misuse of our model for deceptive purposes and advocate for its transparent and responsible use. 
Environmental impacts due to the high computational requirements of LLMs are also considered. Our model aims to reduce training resources, and we encourage sustainable practices in AI research.
% Finally, we are mindful of the impacts on the labor market and commit to engaging with communities to ensure that our technology supports rather than displaces workers. We call for ongoing ethical engagement from the research community to ensure responsible advancement in language technologies.
% \bibliographystyle{plain} % 或其他的参考文献风格
% \bibliography{acl_latex} % 这里替换为你的.bib文件名，不包括后缀
% \section*{Acknowledgements}
% % Entries for the entire Anthology, followed by custom entries
\bibliography{acl_latex}

\appendix

\section{Appendix A}
\label{sec:appendix}

In this section, we provide the detailed information about translation results on Qwen-1b8 and Qwen-7b models. The results are shown in Table~\ref{table3} and Table~\ref{table4}.
% \usepackage{multirow}
\begin{table*}[]
\resizebox{2.1\columnwidth}{!}{
\begin{tabular}{c|cccccccl|ccllllll}
\hline
{Model}      & \multicolumn{8}{c|}{En-X}                            & \multicolumn{8}{c}{X-EN}                                                                                                                       \\ \cline{2-17} 
                            & Ar  & De   & Es   & Fr   & Ru   & Ko   & Zh   & Ja   & Ar   & De   & \multicolumn{1}{c}{Es} & \multicolumn{1}{c}{Fr} & \multicolumn{1}{c}{Ru} & \multicolumn{1}{c}{Ko} & \multicolumn{1}{c}{Zh} & Ja  \\ \hline
Qwen-1b8                    & 1.4 & 16.9 & 22.2 & 22.4 & 9.4  & 4.0    & 35.6 & 8.8  & 15.9 & 30.7 & 28.2                   & 30.8                   & 25.8                   & 6.2                    & 22.1                   & 8.0   \\
Qwen-1b8-CT                 & 7.3 & 21.3 & 25.1 & 26.7 & 12.4 & 11.9 & 27.2 & 20.5 & 24.4 & 31.0   & 28.9                   & 31.0                     & 23.2                   & 6.5                    & 15.2                   & 7.2 \\
Qwen-1b8-MoE-CT (Experts=1) & 7.1 & 20.7 & 24.9 & 26.0   & 11.8 & 12.6 & 29.7 & 18.8 & 22.8 & 31.3 & 28.5                   & 30.8                   & 24.0                     & 6.0                      & 17.8                   & 7.9 \\
Qwen-1b8-MoE-CT (Experts=2) & 7.5 & 21.0   & 25.3 & 26.5 & 12.3 & 12.1 & 30.8 & 20.2 & 24.8 & 31.7 & 29.2                   & 31.2                   & 26.1                   & 6.5                    & 18.5                   & 7.4 \\ \hline
\end{tabular}}
\caption{Translation results on Qwen-1b8 in details.}
\label{table3}
\end{table*}

% \usepackage{multirow}
\begin{table*}[]
\resizebox{2.1\columnwidth}{!}{
\begin{tabular}{c|cccccccc|cccccccc}
\hline
{Model}                          & \multicolumn{8}{c|}{En-X}                                                                                                                                                                                  & \multicolumn{8}{c}{X-EN}                                                                                                                                                                                  \\ \cline{2-17} 
                                                & Ar                   & De                       & Es                       & Fr                       & Ru                       & Ko                   & Zh                       & Ja                    & Ar                   & De                       & Es                       & Fr                       & Ru                       & Ko                   & Zh                       & Ja                   \\ \hline
Qwen-7b                                         & 7.1                  & 27.7                     & 30.6                     & 35.9                     & 16.9                     & 13.4                 & 40.1                     & 20.0                    & 31.5                 & 41.3                     & 34.7                     & 39.6                     & 34.3                     & 14.0                   & 29.0                       & 17.2                 \\
Qwen-7b-CT                                      & 11.7                 & 30.0                       & 32.2                     & 36.3                     & 19.1                     & 21.0                   & 37.7                     & 31.6                  & 35.2                 & 40.9                     & 34.4                     & 39.0                       & 33.6                     & 15.8                 & 26.8                     & 16.3                 \\
Qwen-7b-MoE-CT (Experts=2)                      & 11.6                 & 30.0                       & 32.3                     & 36.8                     & 18.9                     & 21.7                 & 39.3                     & 31.9                  & 35.4                 & 41.1                     & 34.5                     & 39.3                     & 34.0                       & 14.3                 & 27.5                     & 16.6                 \\
Qwen-7b-MoE-CT (Experts=4)                      & 11.6                 & 30.6                     & 32.7                     & 37                       & 19.4                     & 23.0                   & 39.8                     & 32.2                  & 34.7                 & 41.2                     & 34.4                     & 39.5                     & 34.0                       & 14.9                 & 27.7                     & 17.6                 \\
\multicolumn{1}{l|}{Qwen-7b-MoE-CT (Experts=8)} & \multicolumn{1}{l}{11.8} & \multicolumn{1}{l}{30.5} & \multicolumn{1}{l}{32.8} & \multicolumn{1}{l}{37.2} & \multicolumn{1}{l}{19.3} & \multicolumn{1}{l}{23.5} & \multicolumn{1}{l}{39.8} & \multicolumn{1}{l|}{32.5} & \multicolumn{1}{l}{35.4} & \multicolumn{1}{l}{41.1} & \multicolumn{1}{l}{34.4} & \multicolumn{1}{l}{39.2} & \multicolumn{1}{l}{34.1} & \multicolumn{1}{l}{15.5} & \multicolumn{1}{l}{28.1} & \multicolumn{1}{l}{18.0} \\ \hline
\end{tabular}}
\caption{Translation results on Qwen-7b in details.}
\label{table4}
\end{table*}

\end{document}